\documentclass[11pt]{article}

\usepackage[final]{latex/acl}

\usepackage{times}
\usepackage{latexsym}
\usepackage{bm}
\usepackage{amsmath}
\usepackage{amsfonts}
\usepackage{array}
\usepackage{booktabs} 
\usepackage{multirow}
\usepackage{hyperref}
\usepackage{xurl}     
\usepackage{tcolorbox}
\usepackage{float}
\usepackage{enumitem}

\usepackage[T1]{fontenc}

\usepackage[utf8]{inputenc}

\usepackage{microtype}

\usepackage{inconsolata}

\usepackage{graphicx}

\title{Reasoning-Aware AIGC Detection via Alignment and Reinforcement}

\author{
 \textbf{Zhao Wang\textsuperscript{1}\thanks{Equal contribution.}},
 \textbf{Max Xiong\textsuperscript{2}\footnotemark[1]},
 \textbf{Jianxun Lian\textsuperscript{3}},
 \textbf{Zhicheng Dou\textsuperscript{1}}
\\
 \textsuperscript{1}Gaoling School of Artificial Intelligence, Renmin University of China, \\
 \textsuperscript{2}Duke University,
 \textsuperscript{3}Microsoft Research Asia, 
\\
 \texttt{lilin22wz@gmail.com},
 \texttt{jianxun.lian@outlook.com},
 \texttt{dou@ruc.edu.cn}
}

\begin{document}
\maketitle
\begin{abstract}
The rapid advancement and widespread adoption of Large Language Models (LLMs) have elevated the need for reliable AI-generated content (AIGC) detection, which remains challenging as models evolve. We introduce AIGC-text-bank, a comprehensive multi-domain dataset with diverse LLM sources and authorship scenarios, and propose REVEAL, a detection framework that generates interpretable reasoning chains before classification. Our approach uses a two-stage training strategy: supervised fine-tuning to establish reasoning capabilities, followed by reinforcement learning to improve accuracy, improve logical consistency, and reduce hallucinations. Extensive experiments show that REVEAL achieves state-of-the-art performance across multiple benchmarks, offering a robust and transparent solution for AIGC detection. The project is open-source at \url{https://aka.ms/reveal}
\end{abstract}

\section{Introduction}

The rapid advancement of Large Language Models (LLMs) has ushered in an era where AI-generated content (AIGC) is increasingly pervasive and often indistinguishable from human writing. As models approach human-level fluency and coherence \cite{achiam2023gpt}, the ability to reliably discern machine-authored text becomes critical for maintaining integrity across numerous domains. Beyond academic publishing—where undisclosed AIGC threatens to undermine the authenticity of research papers and peer reviews \cite{perkins2023academic,su2023hc3}—AIGC detection is equally crucial in domains like telecommunications fraud prevention, where malicious actors deploy AI to impersonate humans \cite{ciancaglini2020malicious}. A robust, reliable AIGC detector thus serves as an essential safeguard, enabling verification of authorship and upholding trust in digital communications.

Existing approaches to AIGC detection have largely relied on statistical classifiers \cite{log-likelihood,entropy} or black-box neural models \cite{robert}, which often exploit surface-level patterns and struggle to generalize as LLMs evolve \cite{hc3}. While benchmarks such as M4~\cite{m4} and LOKI~\cite{loki} have broadened the scope of evaluation, their data scale remains limited compared to real-world requirements and often fails to include outputs from the latest state-of-the-art models. In this work, we aim to consolidate and advance the field of AIGC detection by introducing a more comprehensive benchmark and a reasoning-driven detector that generalizes effectively to evolving generative technologies.

To support this goal, we construct \textbf{AIGC-text-bank}, a large-scale, multi-domain dataset that includes authentic human writing, fully machine-generated ({AI-Native}) text, and human-authored text polished by AI ({AI-Polish}). Our corpus is sourced from 10 diverse domains and generated using 12 different LLMs—including the latest proprietary and open-weight models—providing a realistic and challenging testbed for detecting authorship in both pure and hybrid scenarios. Its parallel structure ensures that each human reference is paired with AI-generated counterparts, enabling controlled comparisons and finer-grained analysis.

We further introduce \textbf{REVEAL} (Reasoning-Enhanced Verification and Evaluation for AI Language), a novel framework that shifts detection from opaque classification to transparent, reasoning-based decision-making. REVEAL is trained in two stages: first, supervised fine-tuning (SFT) initializes the model by distilling concise and effective rationales from OpenAI o3, serving as an imitation learning phase; then, reinforcement learning (RL) extends these capabilities by refining reasoning chains to improve logical consistency, reduce hallucinations, and ultimately surpass the teacher model's performance. By explicitly modeling the \emph{Think-then-Answer} process, our approach not only achieves high accuracy but also provides interpretable evidence for each prediction. Extensive experiments across five benchmarks demonstrate that REVEAL outperforms existing black-box detectors and general-purpose LLMs in both binary and fine-grained settings, while maintaining strong generalization under domain shift and adversarial challenges.

In summary, our contributions are threefold:
\begin{itemize}[leftmargin=*]
    \item We construct and will release \textbf{AIGC-text-bank}, a large-scale, multi-domain dataset featuring state-of-the-art LLM outputs, providing a comprehensive training resource as well as a benchmark for AIGC detection research.
    \item We propose \textbf{REVEAL}, a reasoning-driven detection framework that combines SFT and RL to produce accurate and interpretable authorship judgments.
    \item We conduct extensive experiments showing that our method sets a new state of the art in generalization and fine-grained discrimination, providing a trustworthy foundation for real-world AIGC detection.
\end{itemize}

\section{Methodology}

\begin{figure*}[t]
  \centering
  \includegraphics[width=\textwidth]{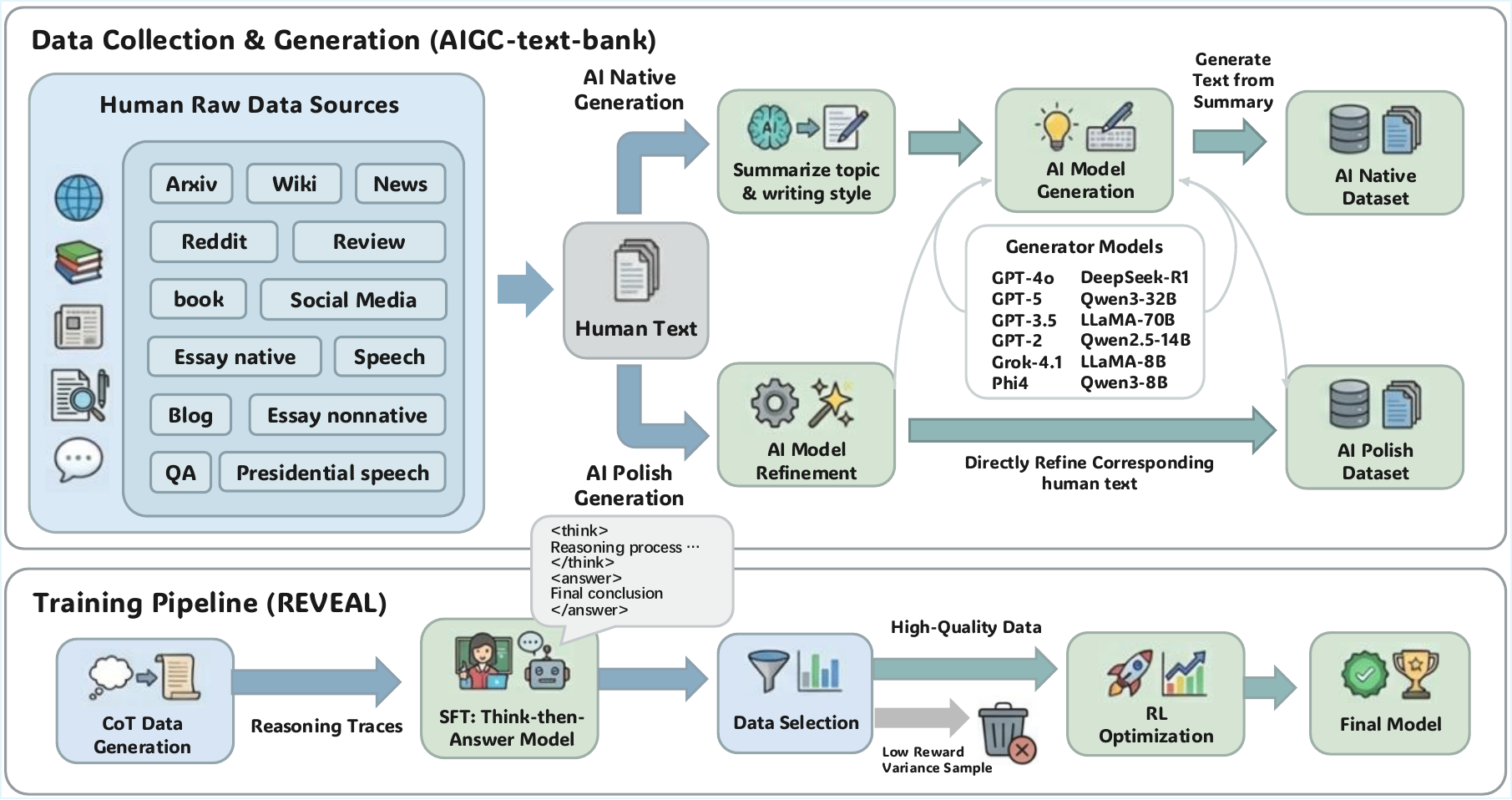}
  \caption{The Overall Framework}
  \label{fig:framework}
\end{figure*}


As LLMs rapidly advance, traditional AIGC detectors relying on superficial statistical cues (e.g., log-likelihood~\cite{log-likelihood}, entropy~\cite{entropy}) become increasingly inadequate, especially in complex, real-world scenarios like AI-Polished content. To address this, we pursue three goals: developing an \textsl{interpretable} LLM-based detector that distinguishes AI from human text with reasoning; extending detection to differentiate \textsl{AI-Native}, \textsl{AI-Polished}, and \textsl{Human content}; and enabling predictive \textsl{uncertainty estimation}. Our methodology constructs a comprehensive, multi-scenario dataset and employs a two-stage training framework of supervised fine-tuning and reinforcement learning to build a detector capable of robust, human-readable reasoning.

\subsection{Dataset Construction}\label{sec:dataset}
\label{ourdataset}

\begin{table}[tp]
\centering
\setlength{\tabcolsep}{5pt}
\caption{Statistics of the AIGC-text-bank dataset.}
\begin{tabular}{l r r c}
\toprule
 & Samples & Total Tokens & \#.LLMs \\
\midrule
Human & 66,979 & 22,535,085 & - \\
AI-Native & 699,052 & 195,392,025 & 12 \\
AI-Polish & 732,248 & 205,644,529 & 12 \\
\bottomrule
\end{tabular}
\label{tab:dataset}
\end{table}

Existing datasets for AIGC detection often suffer from two critical limitations: they fail to include content generated by the latest state-of-the-art LLMs (e.g., GPT-5), and they overlook the nuanced paradigm of human-AI collaborative writing. To bridge this gap, we construct \textbf{AIGC-text-bank}, a comprehensive multi-domain and multi-LLM dataset designed to enhance detector capabilities in real-world human-AI text discrimination. It is structured as a parallel corpus where each human-written document is paired with corresponding AI-generated counterparts. As illustrated in Figure~\ref{fig:framework}, the construction pipeline encompasses three distinct text categories: authentic human writing, fully AI-generated text (\textbf{AI-Native}), and human-authored text refined by AI (\textbf{AI-Polish}).

\paragraph{Human Data Collection}

To establish a robust human baseline, we collect 66,979 authentic human-written documents across 10 diverse domains, including academic papers, social discussions, encyclopedic entries and literature. This diversity ensures extensive coverage of various linguistic styles and structural formats. To mitigate the risk of inadvertently including AI-generated text, we source documents published strictly before the release of ChatGPT (November 30, 2022), thereby establishing a temporal cutoff prior to the widespread public use of advanced, human-like language models. More details about the human subset are provided in the Appendix~\ref{app:human_data}.

\paragraph{Generator Models}

To mitigate architectural inductive biases and capture a broad stylistic spectrum, we employ a diverse ensemble of LLMs varying in parameter scales and performance profiles. Specifically, our generator pool includes state-of-the-art proprietary models (e.g., GPT-5, Grok-4), representative open-source models (e.g., DeepSeek R1, Llama 3.3, Qwen 3, and Phi-4) and legacy models (e.g., GPT-2) to capture the evolutionary trajectory of generative styles. These models serve as the backbone for synthesizing both AI-Native and AI-Polish subsets. We provide the full list of models and details in Appendix~\ref{app:model_config}.

\paragraph{AI-Native Generation}

We propose a semantically aligned reconstruction pipeline to separate intrinsic linguistic signatures from surface-level topical differences. To ensure the synthetic content remains grounded in real-world contexts, we employ GPT-4o to extract structured meta-attributes from the human reference corpus. For each document, GPT-4o distills a concise thematic summary (e.g., topic, key points) and, where pertinent, a profile of its linguistic style (e.g., formal, narrative, conversational). This transformation preserves the domain diversity of the original data while providing a controlled framework for synthesis. Leveraging these meta-attributes, we task the aforementioned 12 generator models with producing content that adheres strictly to the specified topics and writing styles. In this process, we implement two strategic constraints to ensure both the fairness and the complexity of the dataset. First, 
we strictly align the output length with human references to eliminate length as a confounding variable. Furthermore, to simulate real-world scenarios and increase the classification difficulty, we introduce a prompt-based intervention on 20\% of the data samples. In these cases, generators are instructed to imitate human writing styles, making the AI-Native subset both length-matched and more challenging to distinguish stylistically. Detailed data distributions across models are provided in Appendix~\ref{app:ai_nativedata}.

\paragraph{AI-Polish Generation}


To address the realistic and nuanced scenario of human-AI collaborative writing—where, in contexts like academic writing, using AI to polish a human-authored draft is often permissible, whereas generating content directly with AI may constitute a violation of integrity—we introduce the {AI-Polish} subset. This category consists of human-authored texts refined by our generator models to improve fluency and style while strictly preserving the original semantic intent and logical structure. Thus, while the surface presentation may bear AI stylistic signatures, the core ideas remain human-originated, substantively differing from AI-Native content. This subset provides a more challenging detection benchmark, requiring the identification of subtle machine interventions within largely human documents. Examples and data distributions are provided in Appendix~\ref{app:ai_polishdata}.

\subsection{Reasoning Initialization via SFT}
\label{sec:sft}

Conventional detection models treat the task as a discriminative classification problem, often relying on superficial statistical cues or opaque black-box neural models. We argue that a robust, generalizable, and trustworthy detector should instead articulate a human-readable reasoning process before making a decision, rendering the classification transparent and grounded. To this end, we adopt a Think-then-Answer paradigm, which requires the model to base its final verdict on explicit reasoning and concrete evidence extracted from the text.

Since the dataset constructed in Section~\ref{sec:dataset} contains only category labels without explanatory rationales, we leverage the advanced reasoning capabilities of a state-of-the-art LLM, {OpenAI o3}, to augment it with high-quality reasoning trajectories. As illustrated in Figure~\ref{fig:framework}, we employ a hindsight analysis strategy: instead of asking the teacher model to predict the label (which could amplify errors), we provide o3 with the input text $\bm{x}$ and its ground-truth label $\bm{y}$. The model is then instructed to reconstruct a plausible decision-making process, explicitly articulating why $\bm{x}$ belongs to category $\bm{y}$. For example, given an AI-polished text, o3 is prompted to pinpoint the subtle tensions between the underlying human-authored logic and the surface-level AI stylistic artifacts.

This yields a reasoning-augmented dataset $\mathcal{D}_{\text{sft}}$. Each training instance is formatted as a sequence $\text{<think>}\bm{r}\text{</think>}\text{<answer>}\bm{y}\text{</answer>}$, where $\bm{r}$ is the generated reasoning trace and $\bm{y}$ is the label. Direct fine-tuning on lengthy reasoning sequences can, however, disperse the model’s attention away from the final prediction. To address this, we employ a \textbf{Outcome-Weighted Objective} that decouples the generation loss:
\begin{equation}
\label{eq:weighted_loss}
\begin{aligned}
    \mathcal{L}_{\text{sft}} = & - \sum_{i=1}^{m} \log P(r_i | \bm{x}, \bm{r}_{<i}) \\
    & - \lambda \sum_{j=1}^{n} \log P(y_j | \bm{x}, \bm{r}, \bm{y}_{<j}),
\end{aligned}
\end{equation}
where $\lambda > 1$ is a coefficient that increases the weight of the answer loss. By emphasizing the second term, the model is encouraged to use the reasoning path $\bm{r}$ as supporting context, while focusing optimization on the final prediction $\bm{y}$.

\subsection{Reasoning Refinement via RL}
\label{sec:rl}

While SFT establishes an initial capacity for generating reasoning chains, the model remains prone to subtle hallucinations, reasoning-answer inconsistencies, and its capabilities are ultimately bounded by those of the teacher model used for data augmentation. To overcome these limitations and further refine the model's reasoning fidelity, we employ RL for direct preference alignment. This stage aims to reduce errors and improve the overall robustness and logical consistency of the generated reasoning.

To maximize the efficiency of RL training, we first construct a high-quality training set through variance-based data selection. After SFT, the model can confidently classify most samples in $\mathcal{D}_{\text{sft}}$, which provide minimal learning signal. We therefore focus on uncertain, borderline cases where the model's predictions are inconsistent. For each prompt $x$, we perform $K$ stochastic rollouts using the SFT model and compute a binary correctness score $s_k \in \{0, 1\}$ for each rollout. The RL dataset $\mathcal{D}_{\text{rl}}$ is then constructed by selecting only those samples where the model exhibits prediction variance:
\begin{equation}
    \mathcal{D}_{\text{rl}} = \left\{ x \in \mathcal{D}_{\text{sft}} \;\middle|\; 0 < \sum_{k=1}^K s_k(x) < K \right\}.
\end{equation}
By excluding samples with deterministic success or failure, this filtering ensures RL training concentrates on high-uncertainty instances, thereby maximizing the informative gradient signal.

For policy optimization, we adopt the DAPO algorithm~\cite{dapo} which employs decoupled clipping thresholds to independently constrain policy updates from below and above, effectively preventing entropy collapse and promoting stable convergence. The objective is to maximize the expected return over groups of $G$ sampled outputs:
\begin{equation}
    \mathcal{J}(\theta) = \mathbb{E}_{\substack{x \sim \mathcal{D}_{\text{rl}} \\ \{r_i\} \sim \pi_{\theta_{\text{old}}}}} \left( \frac{1}{G} \sum_{i=1}^G \mathcal{L}_i \right),
\end{equation}
where $\mathcal{L}_i$ is the decoupled clipped objective for the $i$-th sample:
\begin{equation}
    \mathcal{L}_i = \min\Big( \rho_i \hat{A}_i, \text{clip}\big(\rho_i, 1-\epsilon_{l}, 1+\epsilon_{h}\big) \hat{A}_i \Big).
\end{equation}
Here, $\rho_i = \pi_\theta(r_i|x) / \pi_{\theta_{\text{old}}}(r_i|x)$ is the importance sampling ratio, and $\epsilon_{l}$, $\epsilon_{h}$ are lower and upper clipping hyperparameters. Following GRPO, the advantage $\hat{A}_i$ is computed by normalizing rewards within the sampled group, eliminating the need for a separate value function:
\begin{equation}
    \hat{A}_i = \frac{R(r_i, y) - \mu_R}{\sigma_R},
\end{equation}
with $\mu_R$ and $\sigma_R$ being the mean and standard deviation of rewards $\{R(r_j, y)\}_{j=1}^G$ for the group.

To effectively guide policy learning, we design a composite reward function $R(r, y)$ that balances final answer accuracy with the structural and logical quality of the reasoning chain:
\begin{equation}
    R(r, y) = R_{\text{acc}}(r, y) + R_{\text{fmt}}(r) + R_{\text{cons}}(r, y).
\end{equation}
The component $R_{\text{acc}} \in \{0, 1\}$ provides a primary outcome signal, awarding $+1$ for a correct final prediction. $R_{\text{fmt}}$ acts as a hard constraint on output format, imposing a penalty of $-1$ if the required output structure is violated. Finally, $R_{\text{cons}}$ is a fine-grained score provided by GPT-4o that evaluates both the internal logical coherence of the reasoning chain and its consistency with the final prediction, preventing the model from exploiting format rewards without genuine reasoning.

\section{Experimental Setup}

We design experiments to answer three core research questions:

\textbf{RQ1:} How does our reasoning-driven detector compare to traditional black-box methods and state-of-the-art LLMs on standard benchmarks?

\textbf{RQ2:} Can our model generalize robustly to unseen downstream detection tasks, particularly those featuring new or evolving label taxonomies?

\textbf{RQ3:} What is the contribution of each key component---the two-stage training, the weighted loss in SFT, and the data selection in RL---to the overall performance?

\subsection{Dataset and Metrics}
We use five diverse benchmarks to verify the accuracy and generalization capabilities of models:

(1) \textbf{AIGC-bench}: As detailed in Section~\ref{ourdataset}, we utilize the held-out test set of our proposed dataset as one benchmark.

(2) \textbf{DetectRL}~\cite{detectrl}: A benchmark that includes multiple specific attack methods (e.g., perturbation attacks), which allows us to assess whether our reasoning framework maintains stability under malicious interference.

(3) \textbf{M4}~\cite{m4}: A large-scale multi-generator corpus covering diverse sources like Wikipedia and Reddit, serving as a standard baseline for distributional generalization.
          
(4) \textbf{Pan}~\cite{pan}: Focuses on human-AI collaboration with a fine-grained 6-class taxonomy (e.g., \textit{Human-written then Machine-polished}), representing complex mixed authorship.

(5) \textbf{LOKI}~\cite{loki}: A comprehensive benchmark encompassing broad text domains (e.g., news, creative writing) designed to evaluate detection capabilities in real-world scenarios.

For these datasets, we primarily use \textsl{Accuracy} and \textsl{Macro F1} as the metrics.

\subsection{Tasks}
To answer the research questions, we design two type of distinct tasks:

\paragraph{Task I: General Detection} 

This task evaluates the model's reasoning performance under two protocols: (1) \textbf{Binary Classification:} For M4, LOKI, DetectRL, Pan, and AIGC-bench, we unify the label spaces into \textit{Human} vs. \textit{AI} to benchmark generalized detection capabilities; and (2) \textbf{Fine-grained Reasoning:} Exclusively on AIGC-bench, we conduct a 3-class classification task (\textit{Human}, \textit{AI-Native}, \textit{AI-Polished}) to verify the model's sensitivity to subtle polishing artifacts.

\paragraph{Task II: Transfer Learning} 

To evaluate our model's potential as a foundation for complex tasks, we initialize with our pre-trained weights and fine-tune the model on the target benchmarks. We evaluate on three tasks requiring high-level adaptation: (1) \textbf{M4 (Domain Adaptation):} Adapting the model to the specific distributions of the multi-generator M4 corpus for robust binary detection; (2) \textbf{DetectRL (Attack Identification):} Distinguishing between specific adversarial attack types (e.g., paraphrasing); and (3) \textbf{Pan (Collaborative Analysis):} Classifying the precise 6-class human-AI collaborative patterns.

\begin{table*}[t]
\centering
\setlength{\tabcolsep}{2.3pt}
\caption{Overall performance comparison on different benchmarks. The best results are in \textbf{bold} and the second are \underline{underlined}.}
\begin{tabular}{>{\raggedright\arraybackslash}p{0.18\linewidth} cc cccccccc cc}
\toprule
\multirow{3}{*}{\textbf{Method}} & \multicolumn{2}{c}{\textbf{In-Domain}} & \multicolumn{8}{c}{\textbf{Out of Domain}} & \multicolumn{2}{c}{\multirow{2}{*}{\textbf{Avg.}}}\\
\cmidrule(lr){2-3} \cmidrule(lr){4-11}
& \multicolumn{2}{c}{AIGC-bench} & \multicolumn{2}{c}{DetectRL} & \multicolumn{2}{c}{M4} & \multicolumn{2}{c}{Pan} & \multicolumn{2}{c}{LOKI} & \\
\cmidrule(lr){2-3} \cmidrule(lr){4-5} \cmidrule(lr){6-7} \cmidrule(lr){8-9} \cmidrule(lr){10-11} \cmidrule(lr){12-13}
& Acc & F1 & Acc & F1 & Acc & F1 & Acc & F1 & Acc & F1 & Acc & F1 \\
\midrule
\multicolumn{13}{l}{\textsc{Discriminative Baseline}} \\
RoBERTa-SFT & \textbf{97.80} & \textbf{97.80} & 93.50 & 93.50 & 73.10 & 72.25 & 85.50 & 85.50 & \textbf{95.70} & \underline{91.94} & \underline{89.12} & \underline{88.20} \\
ImBD & 74.80 & 74.47 & 86.20 & 86.07 & 73.80 & 73.29 & 86.00 & 85.88 & 85.50 & 76.40 & 81.26 & 79.22 \\
Binoculars & 67.60 & 67.17 & 88.60 & 88.60 & 85.60 & 85.49 & 87.90 & 87.84 & 71.50 & 62.92 & 80.04 & 78.40 \\
Fast-DetectGPT & 63.90 & 62.81 & 83.90 & 83.88 & 78.10 & 77.96 & 82.60 & 82.48 & 57.80 & 51.75 & 73.26 & 71.78 \\
\midrule
\multicolumn{13}{l}{\textsc{General LLMs}} \\
Llama3.1-8B & 44.59 & 16.22 & 39.78 & 16.78 & 44.58 & 18.29 & 42.51 & 16.39 & 23.80 & 10.32 & 39.05 & 15.60 \\
GPT-4o-mini & 51.95 & 42.24 & 55.47 & 45.39 & 56.84 & 55.55 & 47.66 & 40.46 & 16.60 & 16.06 & 45.70 & 39.94 \\
GPT-4o & 52.34 & 46.28 & 57.06 & 52.31 & 58.12 & 58.12 & 48.63 & 47.44 & 27.93 & 27.81 & 48.82 & 46.39 \\
GPT-5 & 72.64 & 70.17 & \underline{96.52} & \underline{96.52} & \textbf{89.33} & \textbf{89.31} & 82.57 & 82.57 & 89.45 & 82.18 & 86.10 & 84.15 \\
\midrule
\multicolumn{13}{l}{\textsc{Reasoning LLMs}} \\
Qwen3-8B & 48.90 & 10.45 & 63.86 & 25.14 & 58.84 & 11.80 & 56.34 & 11.31 & 42.17 & 11.19 & 54.02 & 13.98 \\
QwQ-32B & 52.75 & 33.13 & 69.87 & 69.16 & 67.94 & 67.88 & 59.20 & 59.15 & 57.40 & 50.81 & 61.43 & 56.03 \\
OpenAI o3 & 75.81 & 75.06 & 95.74 & 95.73 & \underline{88.03} & \underline{88.03} & \underline{88.59} & \underline{88.58} & 88.24 & 77.92 & 87.28 & 85.06 \\
REVEAL (Ours) & \underline{96.30} & \underline{96.30} & \textbf{97.20} & \textbf{97.20} & 77.86 & 77.25 & \textbf{88.80} & \textbf{88.79} & \underline{95.60} & \textbf{91.98} & \textbf{91.15} & \textbf{90.30} \\
\bottomrule
\end{tabular}
\label{tab:rq1-2}
\end{table*}

\subsection{Baseline}

\noindent \textbf{Discriminative Baseline}: We select supervised RoBERTa-SFT~\cite{robert} and three zero-shot detectors based on token probabilities: Fast-DetectGPT~\cite{fastdetectgpt}, Binoculars~\cite{binoculars}, and ImBD~\cite{imbd}.

\noindent \textbf{General LLMs with Reasoning Prompts}: We evaluate representative proprietary and open-source general LLMs by using Think-then-Answer prompt. We employ GPT-5~\cite{gpt5}, GPT-4o~\cite{gpt4o}, GPT-4o-mini~\cite{gpt4omini} and Llama-3.1-8B-Instruct~\cite{llama} as baselines. 

\noindent \textbf{Reasoning LLMs}: 
This category includes three strong reasoning models: OpenAI o3~\cite{o3}, QwQ-32B~\cite{qwq}, and Qwen3-8B~\cite{qwen3}.

\subsection{Implementation Details}

We implement our framework based on the HuggingFace Transformers~\cite{transformers} and TRL~\cite{trl} Library, using Qwen3-8B as the backbone. For the Imitation Learning stage,  we finetune the model on the constructed 24k reasoning dataset for 3 epochs with a global batch size of 128 and a learning rate of 1e-5. For Preference Alignment, we filter 10k high-uncertainty samples for training; during optimization, we sample 8 outputs per prompt with a temperature of 1.0 and update the policy with a learning rate of 1e-5. For the transfer learning experiments in Setting II, we initialize the model with our aligned weights and apply the same SFT configuration to adapt to downstream benchmarks. All experiments are conducted on four NVIDIA A100-80GB GPUs.

\section{Results}

\subsection{General Detection Performance (RQ1)}

We report experimental results for binary detection (Table~\ref{tab:rq1-2}) and fine-grained classification (Table~\ref{tab:rq1-3}), demonstrating that REVEAL achieves state-of-the-art performance with superior stability. First, REVEAL effectively counters the prediction bias observed in smaller models like Llama 3.1 and Qwen3-8B, which show a large Accuracy–F1 gap by systematically misclassifying fluent AI text as human. Our model closes this gap, achieving metric alignment comparable to larger proprietary models. Second, REVEAL exhibits stronger generalization and robustness: while the supervised RoBERTa-SFT suffers a significant drop on out-of-distribution benchmarks (e.g., from 97.80\% in-domain to 73.10\% on M4), and zero-shot detectors (Fast-DetectGPT, Binoculars, ImBD) struggle to exceed an average accuracy of 81.3\%, our model maintains consistently high cross-domain performance (averaging 91.15\% overall). This indicates that the reasoning-driven paradigm captures more transferable characteristics of AI-generated text.

\begin{table}[t]
\centering
\caption{Fine-grained reasoning results (3-class classification) on AIGC-bench.}
\label{tab:rq1-3}
\setlength{\tabcolsep}{10pt}
\begin{tabular}{>{\raggedright\arraybackslash}p{0.4\linewidth} c c}
\toprule
Method & Acc & F1 \\
\midrule
GPT-5 & 48.30 & 41.00 \\ 
OpenAI o3 & 47.49  & 38.62\\ 
\textbf{REVEAL} (Ours) & \textbf{70.74} & \textbf{70.99}\\ 
\bottomrule
\end{tabular}
\end{table}

\begin{figure}[t]
  \centering
  \includegraphics[width=\columnwidth]{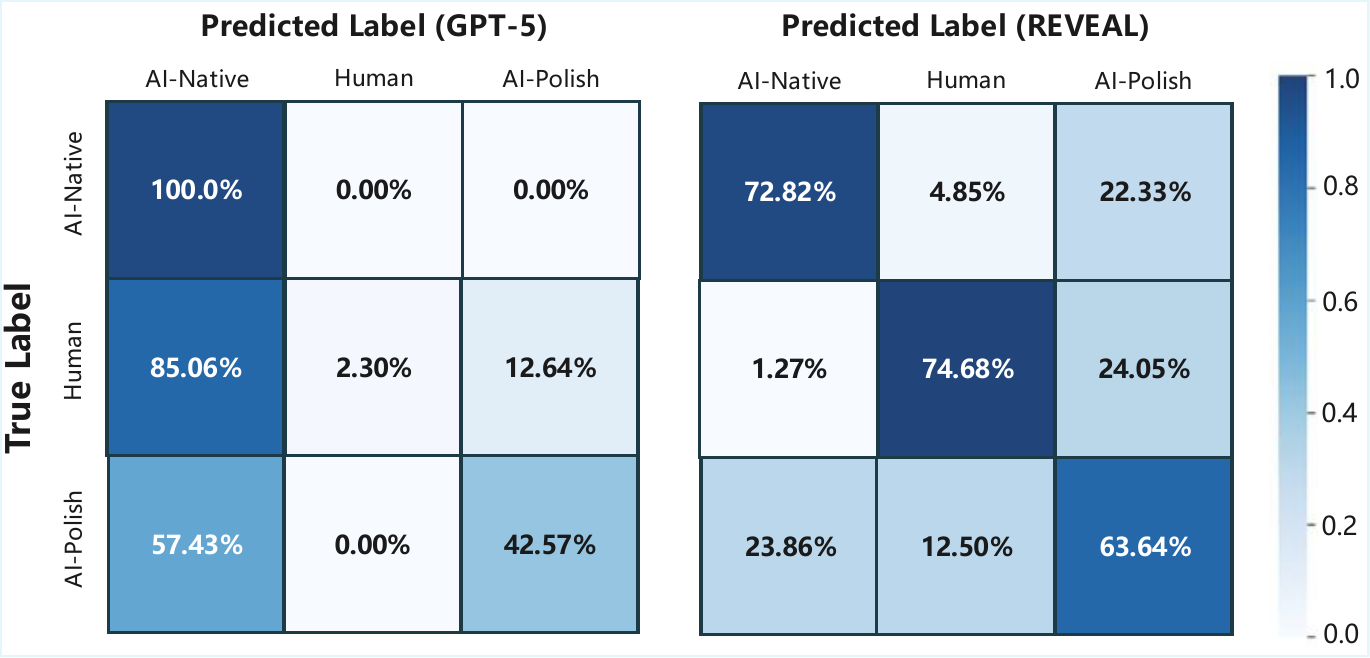}
  \caption{The confusion matrix of GPT-5 and REVEAL.}
  \label{fig:confusion_matrix}
\end{figure}

Distinguishing AI-polished content presents a particular challenge, as shown in Table~\ref{tab:rq1-3}. While strong proprietary models (GPT-5, OpenAI o3) perform near chance (~48\% accuracy), REVEAL attains 70.74\% accuracy. The confusion matrix (Figure~\ref{fig:confusion_matrix}) reveals that GPT-5 exhibits a strong bias towards predicting the AI-Native class, failing to disentangle human logic from AI polish. In contrast, REVEAL demonstrates a more balanced prediction profile, confirming its capability to identify the stylistic artifacts of collaborative writing.

\subsection{Transfer analysis (RQ2)}

\begin{table}[t]
\centering
\setlength{\tabcolsep}{5pt}
\caption{Transfer Learning results.}
\begin{tabular}{>{\raggedright\arraybackslash}p{0.35\linewidth} c c c}
\toprule
\multirow{2}{*}{\textbf{Method}} & M4 & DetectRL & Pan \\
& \small 2 classes & \small 5 classes & \small 6 classes \\
\midrule
Qwen3-8B-SFT & 96.61 & 74.80 & 45.93 \\
OpenAI o3 & 89.53 & 63.69 & 36.75 \\ 
\textbf{REVEAL} (Ours) & \textbf{97.33} & \textbf{75.20} & \textbf{49.07} \\ 
\bottomrule
\end{tabular}
\label{tab:rq2}
\end{table}

Table~\ref{tab:rq2} presents the results of transfer learning experiments, where fine-tuning from REVEAL consistently outperforms initializing from both the standard Qwen3-8B-SFT baseline and the general-purpose OpenAI o3. Across three benchmarks with varying label taxonomies (2, 5, and 6 classes), REVEAL achieves the highest accuracies (e.g., 97.33\% on M4 and 49.07\% on the 6-class Pan dataset). This demonstrates that our model, pretrained with reasoning-driven detection objectives, serves as a superior parameter initialization that effectively transfers to new tasks and complex, unseen label spaces.

\subsection{Ablation Study (RQ3)}

\begin{table}[t]
\centering
\setlength{\tabcolsep}{6pt}
\caption{Performance impact of removing individual components in REVEAL.}
\begin{tabular}{>{\raggedright\arraybackslash}p{0.4\linewidth} c c}
\toprule
Method & In-Domain & OOD Avg. \\
\midrule
\textbf{REVEAL} & \textbf{96.30} & \textbf{89.87} \\
\hspace{1em} w/o SFT & 58.24 & 61.35\\ 
\hspace{1em} w/o RL & 96.16 & 88.69 \\ 
\hspace{1em} w/o Weighted & 93.60 & 87.46\\ 
\hspace{1em} w/o Selection & 96.18 & 89.31\\ 
\hspace{1em} w/o CoT & 91.00 & 85.20 \\ 
\bottomrule
\end{tabular}
\label{tab:rq3}
\end{table}

To evaluate the contribution of each component in our framework, we conduct an ablation study on the setting of binary classification task (the same setting with Table~\ref{tab:rq1-2}):
(1) \textbf{w/o SFT}: removes the whole SFT phase (Section~\ref{sec:sft});
(2) \textbf{w/o RL}: removes the whole RL phase (Section~\ref{sec:rl});
(3) \textbf{w/o Selection}: replaces the uncertainty-based data filtering strategy with random sampling;
(4) \textbf{w/o Weighted}: uses standard next-token prediction instead of the re-weighted loss objective.
(5) \textbf{w/o CoT}: disables the reasoning process, forcing the model to predict the final label directly.

Table~\ref{tab:rq3} presents the results of our ablation study, which validates the necessity of each component. The removal of the SFT phase (\textit{w/o SFT}) severely degrades performance, as the model lacks an initial reasoning structure and struggles to converge efficiently during RL. While SFT establishes the reasoning format, the RL phase is crucial for refining it, as shown by the drop in OOD performance for \textit{w/o RL}. The significant declines observed for \textit{w/o Selection} and \textit{w/o Weighted} highlight the importance of our data and optimization strategies: uncertainty-based filtering forces the model to learn from challenging samples, and the re-weighted loss ensures the final prediction remains the optimization focus. Finally, the substantial drop for \textit{w/o CoT} confirms that explicit reasoning is essential, forcing the model to rely on logical derivation rather than spurious correlations.


\begin{figure}[t]
  \centering
  \includegraphics[width=\columnwidth]{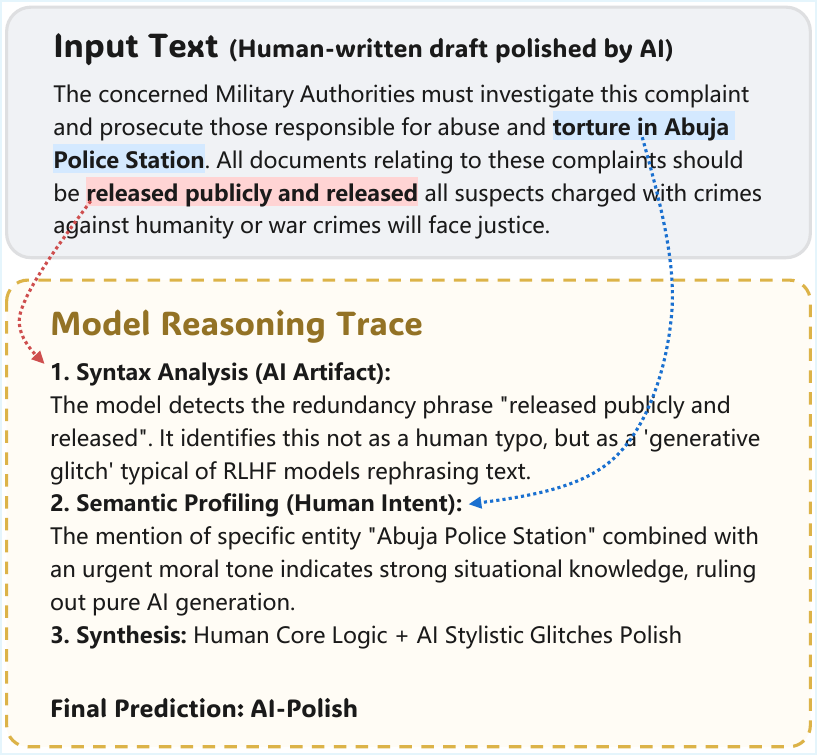}
  \caption{A case study on interpretability in reasoning}
  \label{fig:case1}
\end{figure}

\subsection{Case Study}

\begin{figure*}[t]
  \centering
  \includegraphics[width=\textwidth]{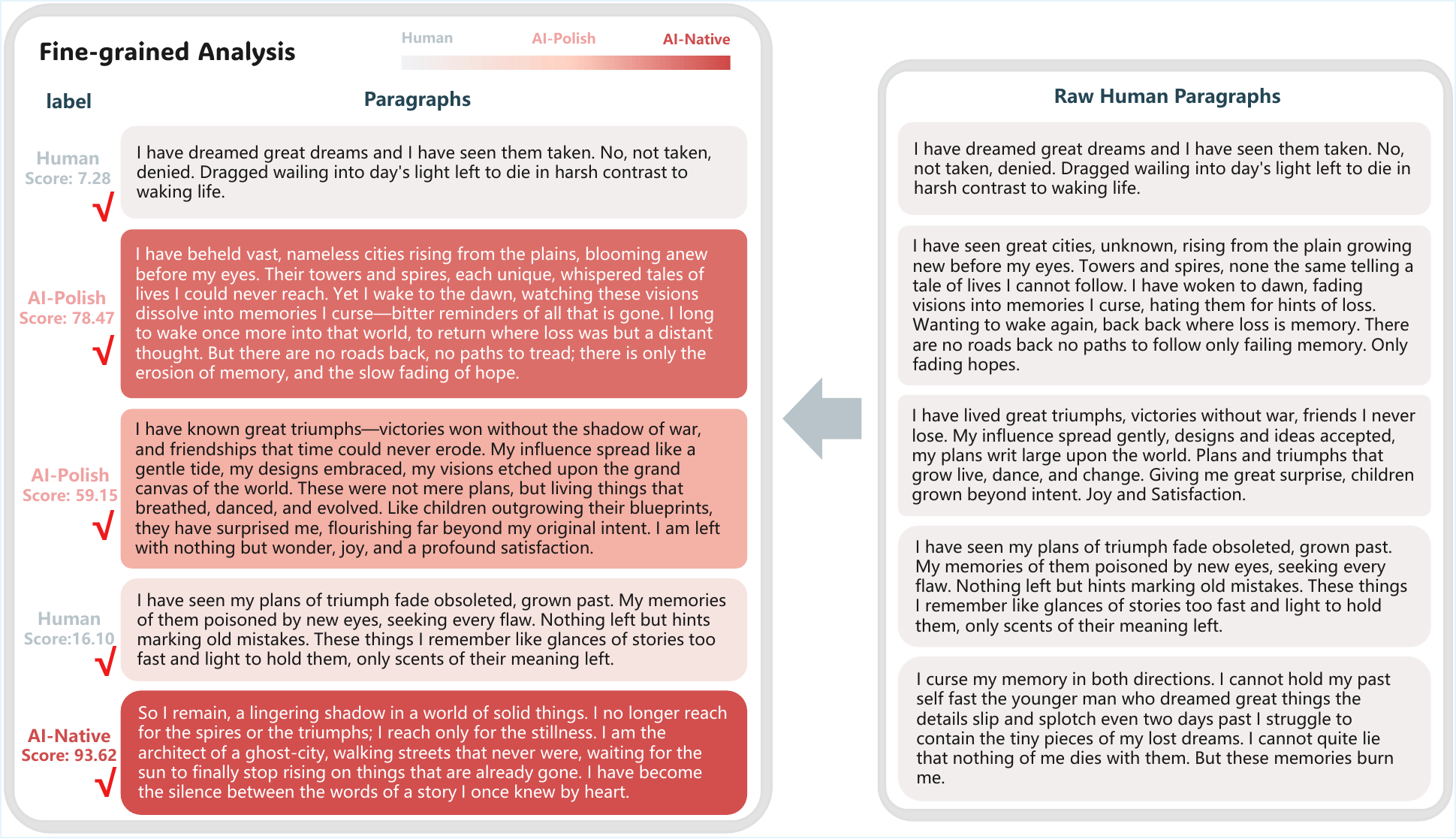}
  \caption{An example on block-wise detection}
  \label{fig:case2}
\end{figure*}

Figure~\ref{fig:case1} illustrates how REVEAL distinguishes complex \textit{AI-Polished} content by reasoning based on explicit linguistic evidence rather than statistic cues. In this case, the model correctly identifies the specific entity ``Abuja Police Station'' and the urgent moral tone as evidence of human authorship. Simultaneously, it flags the redundancy ``released publicly and released'' as a specific generative glitch of AI rephrasing rather than a typo. By weighing these conflicting signals, namely human core logic versus machine syntax, REVEAL derives a transparent and verifiable verdict.

\subsection{Linguistic Analysis}

Based on the reasoning process generated by REVEAL, we conduct a qualitative analysis to identify specific feature sets that distinguish human writing from AI-Native or AI-Polish content.

\paragraph{Human-Written: ``Messy Reality''}
Human text is primarily defined by its spontaneity and lack of standardization. (1) \textbf{Mechanical Irregularities:} Human text frequently contains organic errors, such as comma splices, inconsistent capitalization, and colloquial abbreviations (e.g., "idk", "u"). Such patterns are rarely produced by LLMs without explicit prompting. (2) \textbf{Structural Fluidity:} Human narratives often lack rigid structure, exhibiting meandering thoughts, abrupt topic shifts, or sudden endings without formal conclusions. (3) \textbf{Hyper-Specificity and Emotion:} The Human content also includes unverifiable but vivid details (e.g., specific prices, distinct sensory descriptions) and raw, unfiltered emotions (anger, confusion).

\paragraph{AI-Native: ``Flawless Vacuity''}

AI-Native text is defined by a high degree of polish but a low degree of specific semantic weight. (1) \textbf{Algorithmic Symmetry:} Sentences tend to be balanced in length and rhythm, while grammar and punctuation are invariably perfect. (2) \textbf{Template Adherence:} The text often follows a strict rhetorical structure (e.g., a balanced pros-and-cons list or a standard five-paragraph essay format) and overuses transitional phrases (e.g., "Furthermore," "In conclusion"). (3) \textbf{Generic Content:} The text relies on clichés, safe metaphors, and broad generalizations. Even when hallucinating facts or quotes, the AI tends to generate plausible but fundamentally generic statements that lack idiosyncratic character.

\paragraph{AI-Polished: ``Hybrid''}

AI-Polished text is the most complex, as it combines human intent with algorithmic execution, making it difficult to distinguish. (1) \textbf{The Human Core:} These texts retain the high information density, specific proper nouns, and unique logical leaps of the original human author. The intent remains specific rather than generic. (2) \textbf{The Machine Surface:} Although the content originates from a human author, the syntax is stripped of natural irregularities. The resulting text often exhibits an unusual smoothness relative to the specificity of its content, combining expert-level domain knowledge with the rhythmic uniformity of a language model.

\subsection{Application Discussion}

\begin{figure}[t]
  \centering
  \includegraphics[width=\columnwidth]{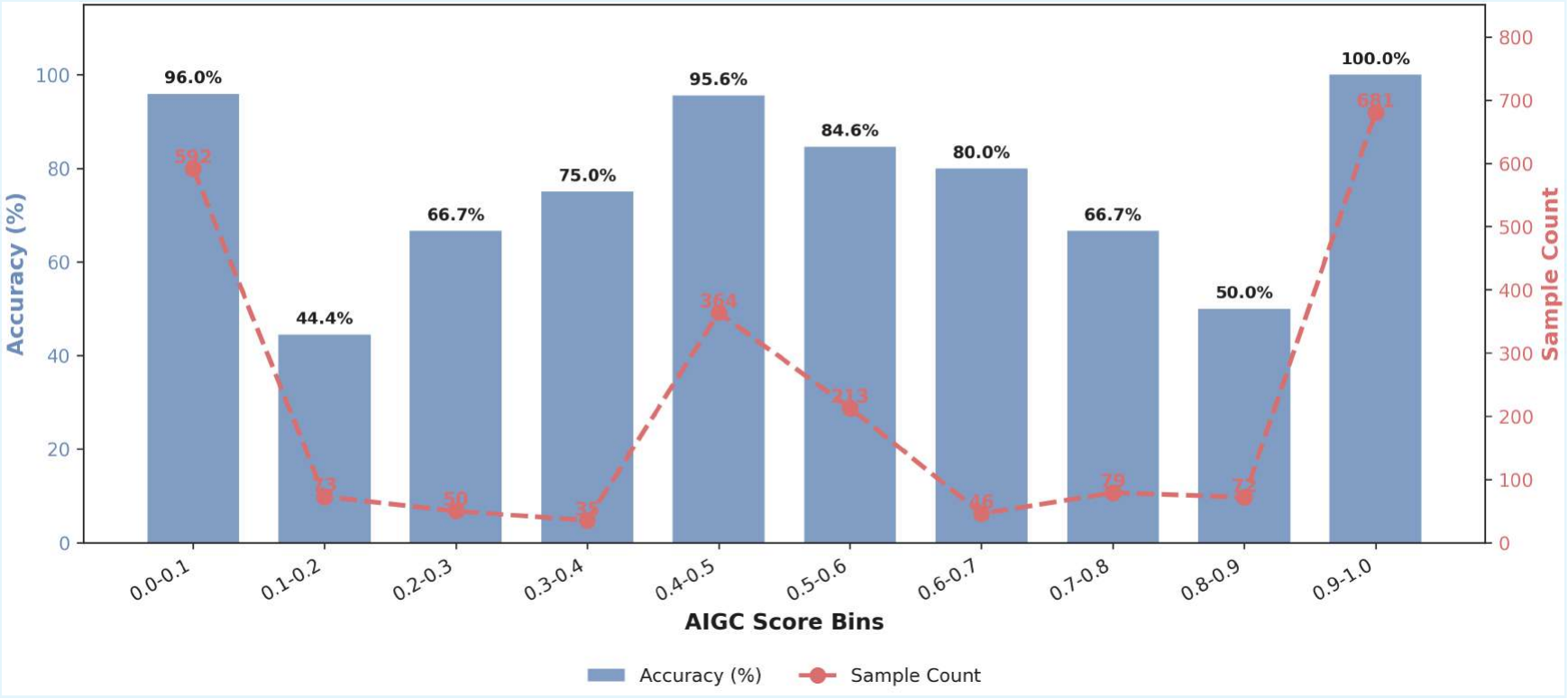}
  \caption{Confidence calibration and correlation with accuracy}
  \label{fig:application}
\end{figure}

Practical applications often require both fine-grained uncertainty estimation and block-wise classification, as lengthy documents may comprise a mixture of human-authored, AI-polished, and AI-generated paragraphs (see Figure~\ref{fig:case2} for an illustrative case). To meet this need, we train another variant  \textbf{REVEAL-Fast} based on AIGC-text-bank. REVEAL-Fast bypasses the reasoning-generation step to output classification results directly. We found that the full REVEAL model, after producing a reasoning trajectory, yields extremely skewed label probabilities (e.g., >99\%), making its output poorly calibrated for confidence estimation. REVEAL-Fast  allows us to derive a well-calibrated ``AIGC score'' by normalizing the logits of the token preceding the final prediction. We map this score such that 0 indicates high confidence in ``Human'' origin, 0.5 in ``AI-Polish'', and 1.0 in ``AI-Native'', with intermediate values reflecting lower certainty. As validated in Figure~\ref{fig:application}, the score exhibits a strong positive correlation with empirical accuracy, confirming its reliability for segmenting documents and assessing the provenance of individual paragraphs with calibrated uncertainty. Further implementation details can be found in Appendix~\ref{app:application}.

\section{Related Works}

\subsection{AIGC Benchmarks and Datasets}
The AIGC benchmarks has shifted from single-source datasets to comprehensive, multi-dimensional evaluations. Early benchmarks like \textsc{TuringBench}~\cite{turingbench} and \textsc{HC3}~\cite{hc3} focused on binary classification across diverse domains. Recently, \textsc{M4} expanded this scope by introducing a multi-generator and multi-lingual corpus to assess detection generalization in the wild~\cite{m4}. To evaluate robustness against adversarial threats, DetectRL constructed datasets simulating real-world scenarios~\cite{detectrl}. Furthermore, \textsc{LOKI} extended detection into the multimodal domain, offering fine-grained annotations for video, image, and text anomalies~\cite{loki}. Despite these advancements, most existing benchmarks treat detection as a document-level binary task, failing to reflect the nature of real-world human–AI writing.

\paragraph{AI-Generated Text Detection.}

Existing detection strategies are generally categorized into white-box and black-box approaches. White-box methods typically require access to the model's internal states or rely on watermarking techniques injected during generation~\cite{watermark}. In contrast, black-box scenarios, which assume access only to the generated text, are more practical for applications. These approaches can be divided into zero-shot methods and supervised classifiers. Zero-shot methods exploit statistical disparities to distinguish AI text from human writing~\cite{detectgpt, fastdetectgpt}. Meanwhile, supervised methods fine-tune Pre-trained Language Models (PLMs) like RoBERTa on large-scale corpora to capture semantic patterns~\cite{log-likelihood}. While these detectors achieve high in-domain accuracy, both zero-shot and supervised methods degrade substantially under real-world adversarial settings, such as AI-polished text or perturbation attacks~\cite{detectrl, krishna2023paraphrasingevadesdetectorsaigenerated}.

\section{Conclusion}
In this work, we construct \textbf{AIGC-text-bank}, a comprehensive dataset for AI-generated content detection, and propose \textbf{REVEAL}, a reasoning-driven framework that replaces black-box classification with interpretable, chain-of-thought analysis. REVEAL is trained in two stages: SFT to initialize reasoning, followed by RL to refine consistency and accuracy. Experiments across five benchmarks show that our approach achieves robust performance with strong generalization. This work bridges high-accuracy detection with human-verifiable explainability, providing a trustworthy foundation for real-world AIGC identification.

\section*{Acknowledgments}
The work was partially done at the Beijing Key Laboratory of Research on Large Models and Intelligent Governance.

\section*{Limitations}

While our work advances AIGC detection through reasoning-driven methods, several limitations merit consideration. 

First, the \textit{Think-then-Answer} paradigm introduces higher inference latency compared to conventional discriminators, posing challenges for real-time applications. Future work may explore model distillation techniques to compress reasoning pathways, parallel processing of reasoning and classification steps, or early-exit mechanisms that adaptively shorten the reasoning chain when confidence is high.

Second, REVEAL currently operates only on textual content and cannot process multimodal inputs such as images, audio, or video. Extending the framework to support multimodal detection would require integrating visual or auditory encoders and designing cross-modal reasoning mechanisms. This direction would allow the model to identify AI-generated content in richer, mixed-modality contexts, better aligning with real-world content consumption.

Finally, the rapid evolution of LLMs presents a persistent challenge, as detectors must continuously adapt to new generator architectures and emerging synthetic patterns. Future research could investigate continual learning strategies that enable detectors to incrementally update with minimal retraining, or develop synthetic data generation pipelines that simulate forthcoming model behaviors. Collaboration with model developers for access to early model outputs could also facilitate more proactive detector adaptation.

\section*{Ethics Statement}

This work aligns with the ACL Code of Ethics. The \textbf{AIGC-text-bank} dataset is curated from publicly available sources (e.g., arXiv, Reddit) in strict compliance with their terms of use, intended only for academic research. We recognize the ethical risks inherent in AIGC detection, particularly the potential for false positives that could result in unjust accusations of misconduct. To address this, our \textbf{REVEAL} framework follows a ``Think-then-Answer'' paradigm, generating interpretable reasoning chains that allow human users to verify evidence rather than relying on opaque automated decisions. We emphasize that this detector should serve strictly as an assistive tool for human oversight, not as an autonomous decision-maker in high-stakes scenarios.

\bibliography{latex/custom}

\appendix

\section{Dataset Construction Details}
\label{app:dataset_details}

In this section, we provide more details about our dataset \textbf{AIGC-text-bank}. Table~\ref{tab:dataset comparison} shows the comparison of our dataset with other AIGC detection datasets. 

\begin{table}[H]
\centering
\setlength{\tabcolsep}{8pt}
\caption{The comparison of AIGC detection dataset.}
\begin{tabular}{l r c}
\toprule
Dataset & Samples & \#.LLMs \\
\midrule
DetectRL & 235,200 & 4 \\
LOKI & 3,359 & 6 \\
M4 & 147,895 & 6 \\
PAN & 361,579 & 3 \\
\textbf{AIGC-text-bank} & \textbf{1,498,279} & \textbf{12} \\
\bottomrule
\end{tabular}
\label{tab:dataset comparison}
\end{table}

\begin{table}[H]
\centering
\setlength{\tabcolsep}{10pt}
\caption{Generator model List. Specific versions include DeepSeek-R1~\cite{deepseek}, Grok 4.1~\cite{grok}, Llama-3.1-8B-Instruct, Llama-3.3-70B-Instruct~\cite{llama}, GPT-5~\cite{gpt5}, GPT-4o~\cite{gpt4o}, GPT-3.5 turbo~\cite{gpt35}, GPT-2 XL~\cite{gpt2}, Phi-4~\cite{phi}, Qwen3-8B, Qwen3-32B~\cite{qwen3}, and Qwen2.5-14B-Instruct~\cite{qwen25}.}
\begin{tabular}{c c}
\toprule
\textbf{Model Family} & \textbf{Specific Version} \\
\midrule
DeepSeek & DeepSeek-R1 \\
\midrule
Grok & Grok 4.1 \\
\midrule
\multirow{2}{*}{Llama} & Llama-3.3-70B-Instruct \\
& Llama-3.1-8B-Instruct \\
\midrule
\multirow{4}{*}{OpenAI} & GPT-5 \\
& GPT-4o \\
& GPT-3.5 turbo \\
& GPT-2 XL \\
\midrule
Phi & Phi-4 \\
\midrule
\multirow{3}{*}{Qwen} & Qwen3-8B \\
& Qwen3-32B \\
& Qwen2.5-14B-Instruct \\
\bottomrule
\end{tabular}
\label{tab:model_params}
\end{table}

\subsection{Human Data Collection}
\label{app:human_data}

\begin{figure}[H]
  \centering
  \includegraphics[width=\columnwidth]{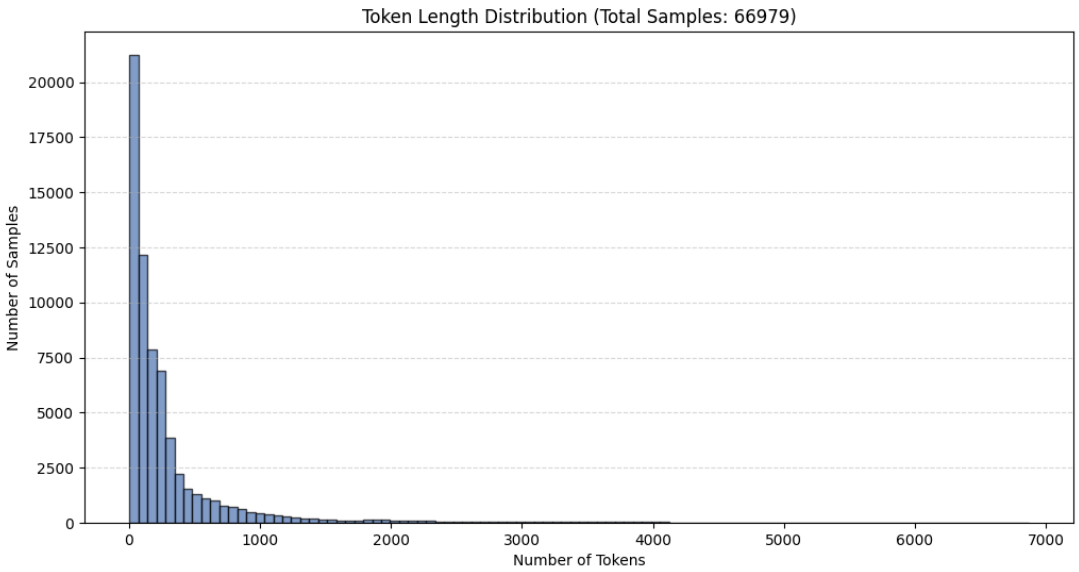}
  \caption{The token distribution of human data.}
  \label{fig:token_human}
\end{figure}

To ensure the diversity and high quality of the human baseline, we curated data from 10 distinct domains. As mentioned in Section~\ref{sec:dataset}, all human texts are published before November 2022 to prevent potential interference from LLM-generated content. Table~\ref{tab:human_sources} presents the detailed statistics and sources for each domain, and Figure~\ref{fig:token_human} illustrates the token distribution of the human data.

\begin{figure}[H]
  \centering
  \includegraphics[width=\columnwidth]{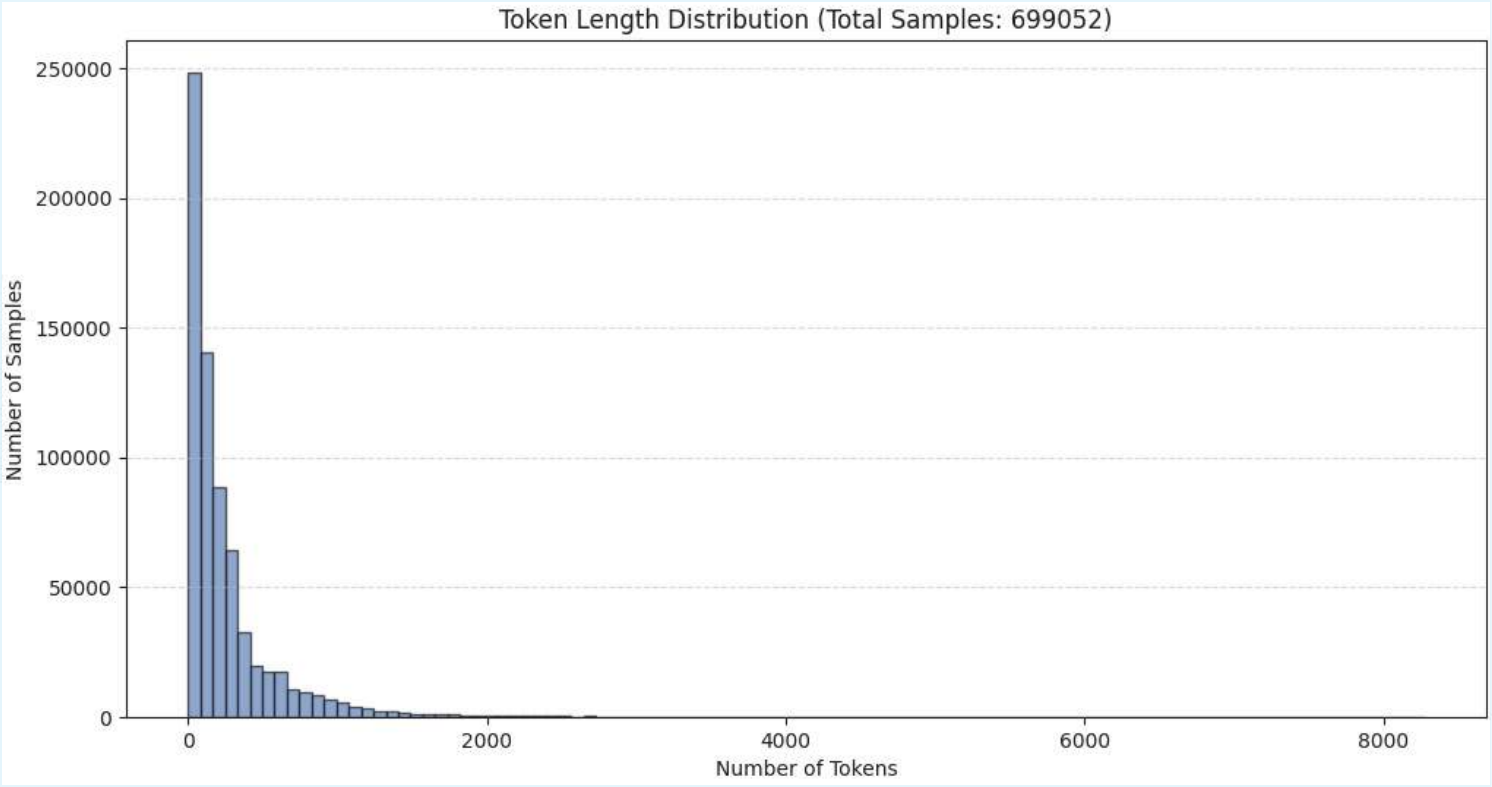}
  \caption{Token distribution of the AI-Native data.}
  \label{fig:token_ai_native}
\end{figure}

\begin{figure}[H]
  \centering
  \includegraphics[width=\columnwidth]{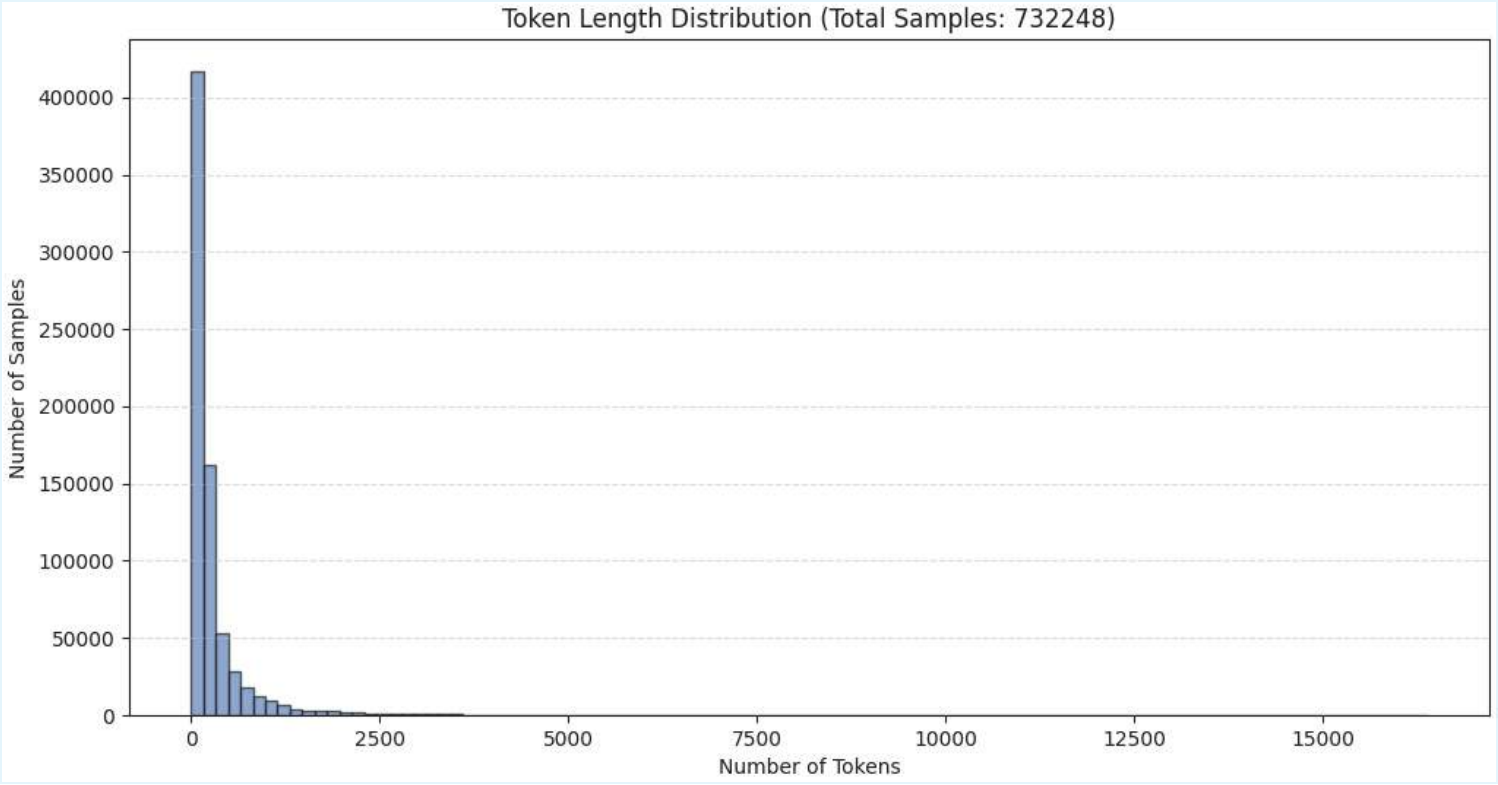}
  \caption{The token distribution of AI-Polish data.}
  \label{fig:token_ai_polish}
\end{figure}

\begin{table*}[t]
\centering
\setlength{\tabcolsep}{5pt}
\caption{Detailed breakdown of Human data sources and statistics across domains.}
\begin{tabular}{l r c p{7cm}}
\toprule
Domain & Samples & Description & Source \\
\midrule
Academic & 9,894 & arXiv papers & \url{https://www.kaggle.com/datasets/Cornell-University/arxiv} \\
\midrule
Blog & 9,986 & Blog posts & \url{https://u.cs.biu.ac.il/~koppel/BlogCorpus.htm}\\
\midrule
Encyclopedic & 558 & Wikipedia & \url{https://www.wikipedia.org/}\\
\midrule
\multirow{2}{*}{Essay} & 1,375 & Native Speaker Essay & \url{https://www.kaggle.com/competitions/llm-detect-ai-generated-text/data} \\
& 3,282 & Non-Native Speaker Essay & \url{https://language.sakura.ne.jp/icnale/} \\
\midrule
Literature & 41 & Classic books & \url{https://www.gutenberg.org/}\\
\midrule
News & 5,000 & News articles & \url{https://huggingface.co/cnn_dailymail/datasets}\\
\midrule
Q\&A & 9,991 & Yahoo Answers & \url{https://www.kaggle.com/datasets/soumikrakshit/yahoo-answers-dataset} \\
\midrule
Reviews & 4,999 & Product reviews & \url{https://huggingface.co/amazon_polarity/datasets}\\
\midrule
\multirow{2}{*}{Social Media} & 9,831 & Twitter posts & \url{https://huggingface.co/datasets/enryu43/twitter100m_tweets} \\
 & 9,446 & Reddit posts & \url{https://developers.reddit.com/docs/capabilities/server/reddit-api}  \\
\midrule
\multirow{2}{*}{Speeches} & 2,450 & TED Talks & \url{https://www.kaggle.com/datasets/rounakbanik/ted-talks} \\
& 126 & Presidential Speech & \url{https://www.kaggle.com/datasets/kboghe/presidentialspeeches} \\
\bottomrule
\end{tabular}
\label{tab:human_sources}
\end{table*}

\subsection{Generator Models}
\label{app:model_config}

We utilize a diverse set of 12 LLMs to generate the AI-Native and AI-Polish subsets. Table~\ref{tab:model_params} summarizes the specific model versions used. 

\subsection{AI-Native Generation} 
\label{app:ai_nativedata}

Table~\ref{tab:ai_native} presents the detailed statistics of the AI-Native subset across 10 domains and 12 LLMs, and Figure~\ref{fig:token_ai_native} illustrates the token length distribution, which closely aligns with the human dataset to minimize length-based bias.

\begin{table*}[t]
\centering
\setlength{\tabcolsep}{3.5pt}
\caption{Detailed statistics of the AI-Native subset.}
\begin{tabular}{l cccccccccc c}
\toprule
Model & Aca. & Blog & Enc. & Essay & Lit. & News & Q\&A & Rev. & Soc. & Spch. & Total\\
\midrule
DeepSeek-R1 & 8539 & 1676 & 20 & 1787 & 0 & 1344 & 1851 & 955 & 3451 & 38 & 19661 \\
Grok 4.1 & 9847 & 9737 & 502 & 4413 & 38 & 4308 & 9591 & 4987 & 19142 & 2526 & 65091 \\ 
Llama-3.3-70B-Instruct & 8077 & 9060 & 543 & 4656 & 28 & 4984 & 8474 & 4375 & 14534 & 1524 & 56255 \\
Llama-3.1-8B-Instruct & 9894 & 9867 & 337 & 4648 & 11 & 4781 & 9978 & 4998 & 19053 & 1043 & 64610 \\
GPT-5 & 9894 & 9532 & 363 & 4635 & 1 & 4780 & 9986 & 4998 & 18879 & 442 & 63510 \\
GPT-4o & 9894 & 9970 & 495 & 4656 & 37 & 4971 & 9989 & 4999 & 19243 & 2264 & 66518 \\
GPT-3.5 turbo & 7344 & 7360 & 386 & 3499 & 9 & 3721 & 7438 & 3758 & 14,217 & 1058 & 65777 \\
GPT-2 XL & 8690 & 7584 & 219 & 4525 & 2 & 4579 & 5269 & 3,295 & 11,372 & 637 & 46172 \\
Phi-4 & 9285 & 9032 & 392 & 4651 & 18 & 4850 & 9285 & 4914 & 18405 & 920 & 62318 \\
Qwen3-8B & 9806 & 9850 & 326 & 4652 & 31 & 4880 & 9905 & 4863 & 18859 & 1719 & 64891 \\
Qwen3-32B & 9891 & 9973 & 554 & 4657 & 35 & 4999 & 9970 & 4998 & 19217 & 2319 & 66613 \\
Qwen2.5-14B-Instruct & 8052 & 9182 & 324 & 4626 & 14 & 4670 & 8976 & 4100 & 16843 & 849 & 57636 \\
\bottomrule
\end{tabular}
\label{tab:ai_native}
\end{table*}

\subsection{AI-Polish Generation}
\label{app:ai_polishdata}

Table~\ref{tab:ai_polish} presents the detailed statistics of the AI-Polish subset, while Figure~\ref{fig:token_ai_polish} illustrates the token length distribution across different samples, providing further insight into its structural characteristics.

\begin{table*}[t]
\centering
\setlength{\tabcolsep}{3.5pt}
\caption{Detailed statistics of the AI-Polish subset.}
\begin{tabular}{l cccccccccc c}
\toprule
Model & Aca. & Blog & Enc. & Essay & Lit. & News & Q\&A & Rev. & Soc. & Spch. & Total\\
\midrule
DeepSeek-R1           & 9824 & 8427 & 141 & 4211 & 2  & 573  & 9009 & 4791 & 18032 & 167  & 55227 \\
Grok 4.1              & 9876 & 9619 & 416 & 4613 & 19 & 4247 & 9757 & 4971 & 19001 & 2336 & 64855 \\
Llama-3.3-70B-Instruct& 9759 & 9691 & 516 & 4634 & 35 & 4763 & 9392 & 4741 & 18541 & 1705 & 63777 \\
Llama-3.1-8B-Instruct & 9892 & 9837 & 536 & 4638 & 37 & 4969 & 9928 & 4998 & 19101 & 1337 & 65273 \\
GPT-5                 & 9888 & 1533 & 97  & 4649 & 5  & 4872 & 9988 & 1017 & 10799 & 358  & 43206 \\
GPT-4o                & 9888 & 9955 & 556 & 4657 & 18 & 5000 & 9955 & 4996 & 19218 & 1572 & 65806 \\
GPT-3.5 turbo         & 9884 & 9977 & 555 & 4657 & 41 & 4997 & 9969 & 4998 & 19250 & 2350 & 66678 \\
GPT-2 XL              & 8261 & 6910 & 82  & 3282 & 2  & 1781 & 6918 & 3627 & 13465 & 94   & 44422 \\
Phi-4                 & 9860 & 9964 & 547 & 4547 & 39 & 4987 & 9962 & 4995 & 19202 & 1563 & 65666 \\
Qwen3-8B              & 9741 & 9898 & 554 & 4348 & 41 & 4990 & 9458 & 4781 & 18929 & 2575 & 65315 \\
Qwen3-32B             & 9886 & 9973 & 557 & 4654 & 41 & 5000 & 9924 & 4979 & 19242 & 2563 & 66819 \\
Qwen2.5-14B-Instruct  & 9852 & 9928 & 545 & 4319 & 41 & 4982 & 9881 & 4935 & 19101 & 1621 & 65204 \\
\bottomrule
\end{tabular}
\label{tab:ai_polish}
\end{table*}

\section{Prompt Engineering}
\label{app:prompts}

In this section, we provide the exact prompt templates used in our framework. We detail the prompts for three distinct stages: (1) Reasoning Data Synthesis (Teacher Model), (2) Standard Detection (REVEAL), and (3) Consistency Evaluation (Reward Model).

\subsection{Reasoning Data Synthesis}
To construct the reasoning-augmented dataset $\mathcal{D}_{\text{sft}}$, we employ a \textbf{hindsight prompting} strategy. Specifically, We provide the teacher model (OpenAI o3) with both the input text and its ground-truth label, instructing it to reconstruct the reasoning process leading to the correct classification. The template is presented in Table~\ref{tab:teacher_prompt}.

\begin{table}[H]
\centering
\small
\begin{tcolorbox}[colback=gray!10, colframe=gray!50, boxrule=0.5pt, arc=2pt, left=4pt, right=4pt, top=4pt, bottom=4pt]
\textbf{Instruction:} A conversation between User and Assistant.\\
The Assistant first thinks in <think>…</think> tags then answers in one word (Human or AI) in <answer>…</answer> tags. \\
Your task: You are given a human-written or AI-generated/edited piece of text. You must determine whether the piece was written/edited by AI or human-written. \\
Let's think step-by-step. Describe inconsistencies/AI artifacts or any clues that this text may be human/written, summarize your analysis, then answer with Human or AI. \\
\\
\textbf{Text:} \{input\_text\}
\end{tcolorbox}
\caption{The Inference Prompt used for REVEAL training and baseline evaluation.}
\label{tab:inference_prompt}
\end{table}

\subsection{Standard Detection Prompt}
For both the Supervised Fine-Tuning (SFT) of REVEAL and the inference evaluation of baseline models, we use a uniform \textbf{Think-then-Answer} prompt. This ensures that the model generates an explicit reasoning chain before predicting the final label. The template is presented in Table~\ref{tab:inference_prompt}.

\subsection{Consistency Reward Prompt}
In the Reinforcement Learning stage, we employ GPT-4o as a reward model to assess the logical consistency of the generated reasoning chain ($R_{\text{cons}}$). To guide its evaluation, we provide the model with \textbf{2 examples} to illustrate how to evaluate the reasoning process. The evaluation template is presented in Table~\ref{tab:reward_prompt}.

\begin{table}[t]
\centering
\small
\begin{tcolorbox}[colback=gray!10, colframe=gray!50, boxrule=0.5pt, arc=2pt, left=4pt, right=4pt, top=4pt, bottom=4pt]
\textbf{Instruction:} You're a forensic writing analyst trained to detect whether a piece of text was written by a human or generated by AI.\\
\\
Below is a passage of text and a known label indicating whether it is Human-written or AI-generated. \\
\\
Your job is to:\\
1. Analyze the text step by step.\\
2. Identify concrete evidence that supports the given label.\\
3. Contrast it with why the opposite label is less likely.\\
4. Write your reasoning in natural language inside <think> tags.\\
5. Conclude with the final label (One word: "Human" or "AI") in <answer> tags.\\
6. Do not use any other tags or formatting.\\
7. Do not explicitly mention the ground truth label in your reasoning. Assume you do not yet know the label.\\
\\
Always ground your analysis in specific stylistic, structural, or semantic features of the text. Avoid generic summaries or descriptions.\\
\\
\textbf{Text:} \{input\_text\}
\\
\textbf{Ground Truth Label:} \{label\} \\
\\
<think>
\end{tcolorbox}
\caption{The Hindsight Prompt used for generating reasoning traces from the Teacher Model (OpenAI o3).}
\label{tab:teacher_prompt}
\end{table}

\begin{table}[t]
\centering
\small
\begin{tcolorbox}[colback=gray!10, colframe=gray!50, boxrule=0.5pt, arc=2pt, left=4pt, right=4pt, top=4pt, bottom=4pt]
\textbf{Instruction:} You are an expert evaluator tasked with assessing whether a model’s final classification is consistent with its reasoning. The model's objective is to determine whether a given piece of text was written by a human or generated by a large language model (LLM). \\
\\
You will be provided with:\\
- An input text: the passage under evaluation.\\
- A reasoning trace, enclosed in <think>...</think> tags, representing the model's chain-of-thought.\\
- A final classification, enclosed in <answer>...</answer> tags, indicating the model’s predicted label (Human or AI).\\
\\
Your task is to return a list of three float scores, each ranging from 0.0 to 1.0, corresponding to the following criteria:\\
1. Answer–Reasoning Alignment: Does the reasoning logically support the final answer? This should be binary (1.0 or 0.0) based on whether the reasoning is consistent with the final classification.\\
2. Groundedness: Is the reasoning grounded in the input text and internally coherent?\\
3. Specificity (Genericness): How specific, informative, and non-generic is the reasoning?\\
\\
Respond strictly with a Python-style list of floats in this format:\\
\text{[alignment\_score, groundedness\_score, genericness]}\\
Do not include any explanations, comments, or extra output.\\
\\
\textbf{Examples:} \{2 examples\}\\
\textbf{Text:}\{original\_text\}\\
\textbf{Model Output:}\{model\_output\}
\end{tcolorbox}
\caption{The Reward Prompt used for evaluating reasoning consistency in RL.}
\label{tab:reward_prompt}
\end{table}

\section{Further Analysis and Discussion}

\subsection{Ablation Study}

To provide deeper insight into the contribution of each component, we visualize the reward curves during the Reinforcement Learning phase. 

\begin{figure*}[t]
  \centering
  \includegraphics[width=\textwidth]{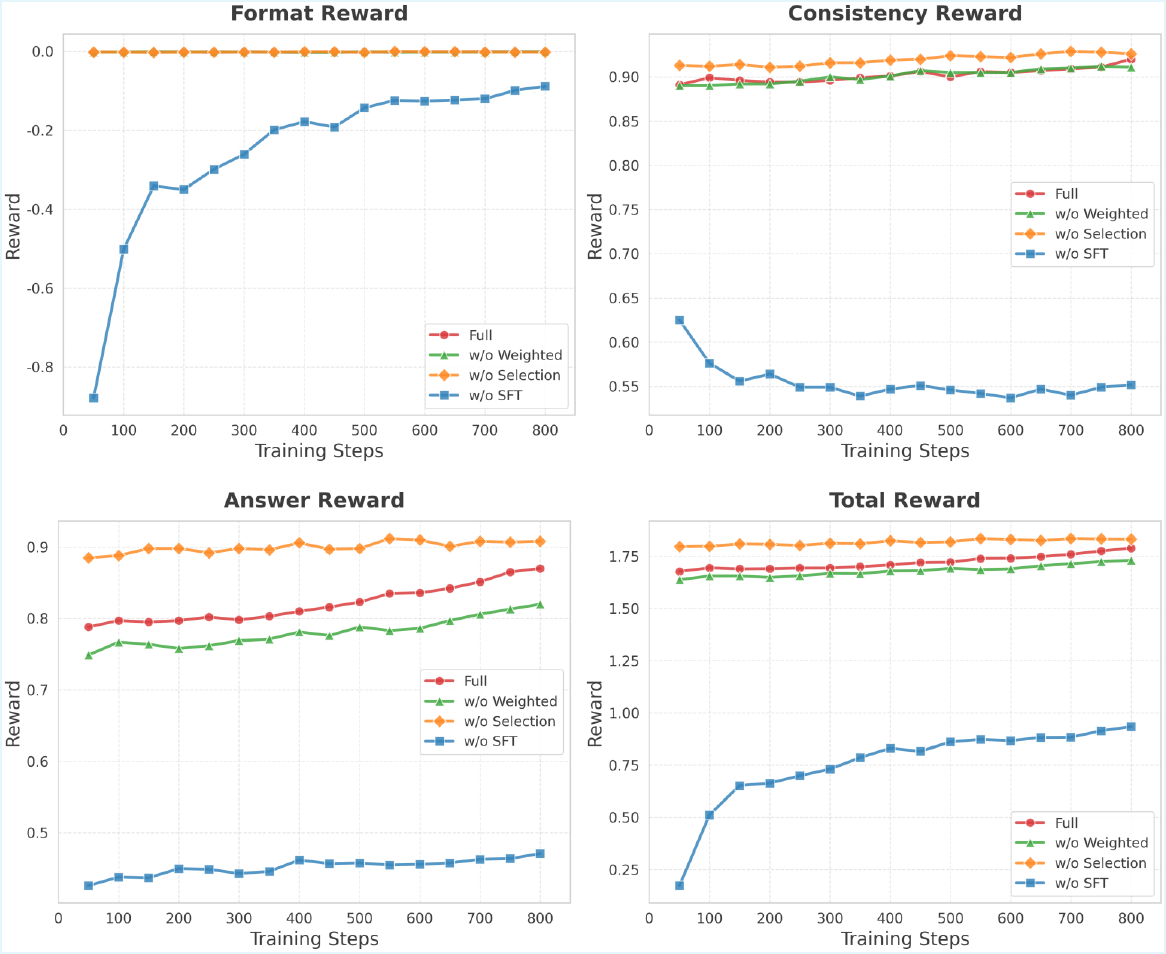}
  \caption{The reward curves during Reinforcement Learning.}
  \label{fig:reward}
\end{figure*}

Figures \ref{fig:reward} display the progression of Total Reward, Answer Reward, Consistency Reward, and Format Reward, respectively. These curves reveal three critical observations regarding the stability and efficiency of our framework:

\paragraph{Impact of Data Selection Strategy.}

As shown in Figure \ref{fig:reward}, the \textit{w/o Selection} variant (orange curve) exhibits slightly higher rewards than the Full model on the validation set, largely because the validation set follows the same distribution as the training data. Random sampling in \textit{w/o Selection} overrepresents high-confidence, statistically abundant easy samples, inflating validation rewards. In contrast, uncertainty-based filtering biases the Full model toward ambiguous, boundary cases. Although this reduces average rewards on an easy validation set, it mitigates overfitting to trivial patterns and leads to substantially improved OOD robustness, as evidenced in Table~\ref{tab:rq3}.

\paragraph{The Necessity of SFT Initialization.}

Removing the SFT stage (\textit{w/o SFT}) leads to a substantial degradation in training efficiency, as reflected by the blue curves. The format reward reveals that, without SFT, the early RL phase is largely consumed by learning basic syntactic structures (e.g., reasoning tags) rather than improving reasoning quality. This cold-start issue propagates to downstream learning: inadequate formatting prevents the model from forming coherent reasoning trajectories, leading to persistently low Consistency Rewards. These results confirm that SFT provides an essential structural prior, enabling the RL phase to concentrate on refining reasoning consistency instead of recovering basic output structure.

\paragraph{Effectiveness of the Full Framework.}
The Full REVEAL model (red curve) demonstrates consistent improvement across all reward dimensions. In contrast to \textit{w/o SFT}, it maintains strong format adherence from the outset, and compared to \textit{w/o Weighted} (green curve), it achieves superior asymptotic performance in both Answer and Consistency rewards. By integrating SFT initialization, weighted loss, and hard-sample mining, the Full model ensures that gains in reward metrics correspond to genuine improvements in reasoning capability.

\subsection{Detailed Application Discussion}
\label{app:application}

While our main model focuses on reasoning, practical scenarios often require fast, fine-grained scanning. To address this, we utilize \textbf{REVEAL-Fast}, trained directly on 3-class labels (Human, AI-Native, AI-Polish) for paragraph-level detection. 

To quantify the model's confidence, we extract the raw logits associated with the final token before the generated label. Let $z_h, z_p, z_n$ represent the output logits for Human, AI-Polish, and AI-Native respectively. We first apply a standard Softmax function to normalize these logits into a probability distribution:

\begin{equation}
P_c = \frac{\exp(z_c)}{\sum_{j} \exp(z_j)}, \quad c \in \{h, p, n\}.
\end{equation}

To map these discrete probabilities onto a continuous spectrum $S \in [0, 1]$, we formulate the score as the mathematical expectation of the ``AI-Generation Degree''. We assign discrete quantization values to each category: 0 for Human, 1 for AI-Native, and 0.5 for AI-Polish. The final AIGC Score $S$ can be calculated as:

\begin{equation}
\begin{aligned}
S &= \mathbb{E}[\text{AI Degree}] \\
&= 0 \cdot P_h + 0.5 \cdot P_p + 1 \cdot P_n \\
&= P_n + 0.5 \cdot P_p
\end{aligned}
\end{equation}

This expectation-based formulation is theoretically sound as it accounts for the inherent uncertainty between categories. For instance, if the model is uncertain between Human and AI-Polish (e.g., $P_h = 0.5, P_p = 0.5$), the score naturally converges to 0.25, correctly indicating a low-risk segment. Conversely, uncertainty between AI-Polish and AI-Native yields a score around 0.75, flagging high-risk content.

To confirm the validity of this scoring mechanism, we conduct a calibration experiment on the 3-class test set of \textbf{AIGC-bench}. We partition the samples into 10 distinct bins based on their predicted AIGC Score (e.g., $[0.0, 0.1), \dots, [0.9, 1.0]$). For each bin, we calculate the accuracy.

As shown in Figure~\ref{fig:application}, AIGC scores near 0, 0.5, and 1 indicate high confidence in the respective classes, while accuracy is also the highest in these regions. This pattern demonstrates that the AIGC Score offers a fine-grained and well-calibrated measure of the model's output confidence.

To illustrate the model's capabilities in real-world scenarios, we provide a fine-grained detection case study in Figure~\ref{fig:case2}. Starting with a raw human-written text, we segmented it into distinct paragraphs and manually constructed a hybrid test case: some paragraphs were left as original human text, others polished by an LLM to simulate AI-Polish content, and the concluding section fully generated to represent AI-Native content. As illustrated in the figure, our model effectively scans the document paragraph-by-paragraph, and provide AIGC score for each section. The visualization highlights the model's ability to differentiate between subtle polishing and complete generation, enabling precise localization of AI content within mixed-source documents.

\end{document}